\documentclass[lettersize,journal]{IEEEtran}
\usepackage{amsmath,amsfonts}
\usepackage{algorithmic}
\usepackage{algorithm}
\usepackage{array}
\usepackage{textcomp}
\usepackage{stfloats}
\usepackage{url}
\usepackage{verbatim}
\usepackage{graphicx}
\usepackage{cite}
\usepackage{subfigure}
\usepackage{hyperref}
\usepackage{color}
\usepackage{dutchcal}
\usepackage{mwe}
\usepackage{multicol}
\usepackage{physics}
\usepackage[T1]{fontenc}
\usepackage{todonotes}
\begin{document}
\title{An Efficient Federated Learning Framework for Training Semantic Communication System}

\author{Loc~X.~Nguyen, Huy~Q.~Le, Ye~Lin~Tun, Pyae~Sone~Aung,
       Yan~Kyaw~Tun,~\IEEEmembership{Member,~IEEE},
Zhu~Han,~\IEEEmembership{Fellow,~IEEE,}~and~Choong~Seon~Hong,~\IEEEmembership{Senior~Member,~IEEE}

\thanks{Loc X. Nguyen, Huy Q. Le, Ye Lin Tun, Pyae Sone Aung, and Choong Seon Hong  are with the Department of Computer Science and Engineering, Kyung Hee University,  Yongin-si, Gyeonggi-do 17104, Rep. of Korea, e-mails:{\{xuanloc088, quanghuy69, yelintun, pyaesoneaung, cshong\}@khu.ac.kr}.}

\thanks{Yan Kyaw Tun is with the Department of Electronic Systems, Aalborg University, 2450 København SV, Denmark, e-mail:{\{ykt\}@es.aau.dk}.
}
\thanks{Zhu Han is with the Electrical and Computer Engineering Department, University of Houston, Houston, TX 77004, and also with the Department of Computer Science and Engineering, Kyung Hee University, Yongin-si, Gyeonggi-do 17104, Rep. of Korea, email:{\{hanzhu22\}}@gmail.com.}

}


\maketitle 

\begin{abstract}
Semantic communication has emerged as a pillar for the next generation of communication systems due to its capabilities in alleviating data redundancy. Most semantic communication systems are built upon advanced deep learning models whose training performance heavily relies on data availability. Existing studies often make unrealistic assumptions of a readily accessible data source, where in practice, data is mainly created on the client side. Due to privacy and security concerns, the transmission of data is restricted, which is necessary for conventional centralized training schemes. To address this challenge, we explore semantic communication in a federated learning (FL) setting that utilizes client data without leaking privacy. Additionally, we design our system to tackle the communication overhead by reducing the quantity of information delivered in each global round. In this way, we can save significant bandwidth for resource-limited devices and reduce overall network traffic. Finally, we introduce a mechanism to aggregate the global model from clients, called FedLol. Extensive simulation results demonstrate the effectiveness of our proposed technique compared to baseline methods.
\end{abstract}

\vspace{-0.1in}
\section{introduction}

\IEEEPARstart{C}onventional communication, which targets accurately sending bit series without understanding its meaning, had approached its bounds by the Shannon capacity. In addition, the growth of connected devices, as well as complex and high-demand applications (e.g., Metaverse and telemedicine) in wireless networks, encourages researchers to look for a new communication paradigm capable of guaranteeing the quality of service (QoS) for all connected devices. Recently, semantic communication, which considers the semantic meaning of transmitted data, has been proposed as a promising pillar for the next generation of communication paradigms. It operates in the second level among three communication levels defined by Weaver \cite{shannon1949mathematical}: bit communication (conventional), semantic communication, and effective communication. The current conventional communication paradigm concentrates on minimizing the bit error rate, while effective communication studies the effectiveness of the receiver behavior corresponding to the message. In this letter, we target the second level problem, particularly on how to exploit privacy-sensitive user data for designing and training semantic communication systems. 

Current works related to semantic communication either focus on improving the system performance by adopting highly advanced deep learning models or resolving different tasks with different kinds of data modality. The work of \cite{farsad2018deep} first proposed a joint source-channel coding for text, and later on \cite{xie2021deep} proposed a Transformer~\cite{vaswani2017attention} model to improve the efficiency of the system for text transmission. Recent advancements in deep learning models have successfully extracted the semantic feature within the image, which has been adopted in semantic communication for classification tasks \cite{kang2022task}. In \cite{zhang2022unified}, a semantic communication system capable of handling diverse tasks with multiple data modalities was proposed. However, none of the aforementioned works addressed data privacy concerns associated with the training process of such semantic communication systems, posing a significant drawback. 


The training process of a semantic communication system is typically carried out in a centralized setting, resulting in costly data collection processes essential to creating a centrally stored training dataset. This approach often overlooks user data privacy, assuming that user data can be exposed to external parties. Given the growing data privacy and security concerns, such a drawback is becoming increasingly unfavorable. Fortunately, a framework called Federated Learning (FL) \cite{mcmahan2017communication} was proposed to take advantage of distributed user data for model training without leaking privacy-sensitive data. The core idea of the framework is to send the learning model to the user sites and train it locally. The trained models from different mobile users are then aggregated at the server through parameter averaging. This entire process constitutes one communication round or global round, which repeats until the model reaches the desired level of performance or a specific termination condition. Nonetheless, FL still faces several critical problems, such as non-independent and identically distributed (non-IID) data, slow convergence, and communication cost \cite{li2020federated1,khan2021federated}. To the best of our knowledge, no prior works attempted to integrate semantic communication with the FL framework in the context of the vision domain, although, \cite{tong2021federated} considered a case study for audio modality.

\textcolor{black}{In this study, we explore the training of semantic communication under the FL framework for image modality, which poses a greater reconstruction challenge compared to audio data}. On top of that, we analyze the complexity of each module in semantic communication, and based on that analysis, we propose a simple yet effective way to reduce the communication overhead in FL, \textcolor{black}{specifically omitting the transmission of channel parameters}. Moreover, we propose a novel aggregation scheme called federated local loss (FedLol) for updating the global model, which offers superior robustness to non-IID scenarios than federated averaging (FedAvg)\cite{mcmahan2017communication}, federated proximal (FedProx) \cite{li2020federated}, and model contrastive federated learning (MOON)\cite{li2021model}.
The main contributions of this letter can be summarized as follows:
\begin{itemize}
    \item By leveraging the FL framework, we harness the private data of mobile users to enhance the performance of semantic communication system, which heavily depends on data availability.
    \item Given the substantial communication overhead associated with each training round in FL, we present an efficient solution to mitigate communication costs by updating a part of the learning model. 
    \item \textcolor{black}{In addition, to improve the performance of the FL framework, we propose a new approach for aggregating the global model, which we refer to as FedLol. It determines each local model’s contribution to the aggregation process based on its local loss at the client. Importantly, this approach does not introduce any additional complex modules or changes to the communication protocol.}
    \item \textcolor{black}{Finally, we conduct thorough simulations to illustrate the effectiveness of the proposed technique in comparison to both traditional and state-of-the-art baseline methods.}
\end{itemize}
\vspace{-0.13in}
\section{preliminary}
\vspace{-0.05in}
\begin{figure*}[t!]
    \centering
    \includegraphics[width=0.70\textwidth]{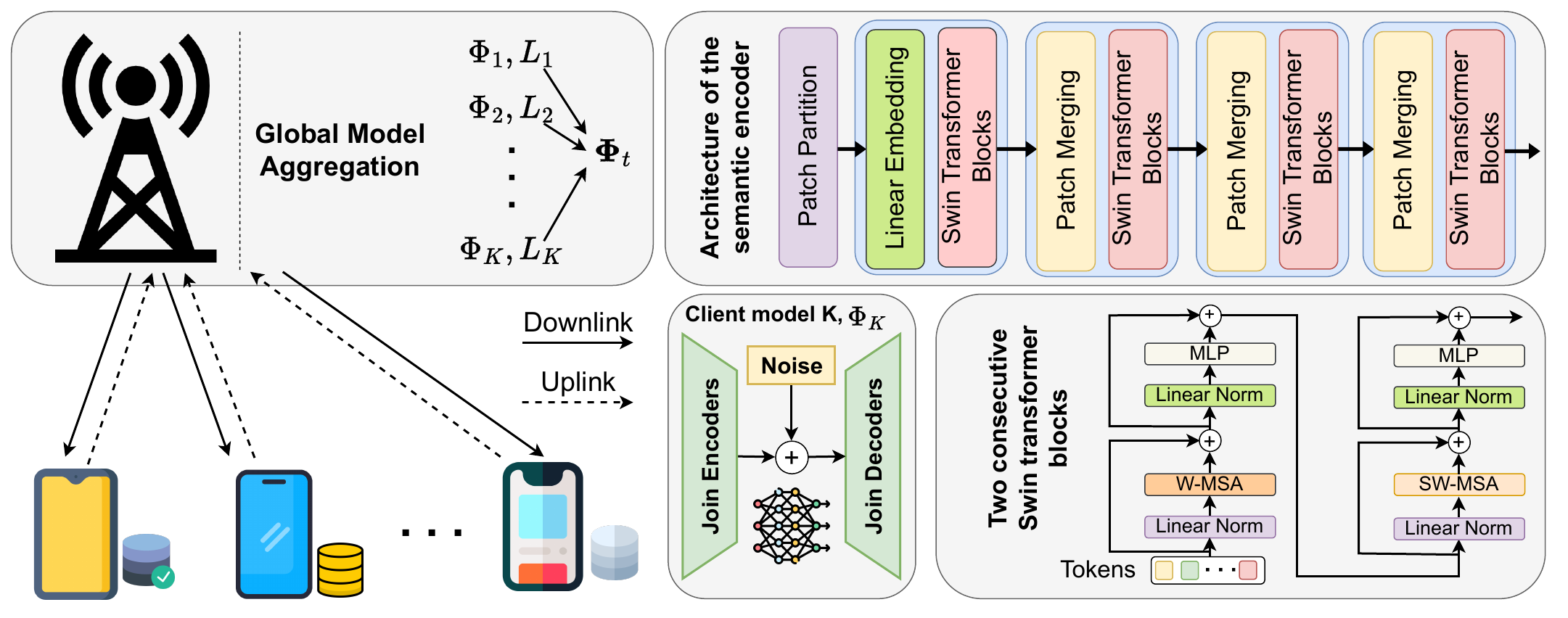}
    \caption{The left-hand side figure demonstrates the proposed Federated Learning framework in a semantic communication system. The right-hand side figure shows the architecture of the Semantic Encoder and the detailed components of the Swin Transformer blocks.}
    \label{systemmodel}
\end{figure*}
The considered system model consists of a centralized server and a set $\mathcal{K}$ of $K$ mobile users, which operate as FL agents as shown in Fig.~\ref{systemmodel}.  
\vspace{-0.13in}
\subsection{Semantic Communication}
Following the common framework of semantic communication, the proposed system consists of four basic modules: \emph{semantic encoder, semantic decoder, channel encoder, and channel decoder}. The semantic encoder and channel encoder will be activated for transmitting, while the other modules are responsible for receiving the upcoming signal. In this letter, we consider the image transmission and reconstruction scenario, which is the most standard task in semantic communication due to its massive dimensional space. The main features of the image will be extracted by the semantic encoder as follows: 
\vspace{-0.01in}
\begin{equation}
    F_{I}= E^{S}_{\alpha}(I),  \forall {I} \in R^{ 3 \times H \times W},
    \vspace{-0.01in}
\end{equation}
where $I$ denotes the image, $R$ is the dimension of the image, $E^{S}$ represents the semantic encoder, which is implemented as a deep learning model, and $\alpha$ is the associated learning parameter of the model. The semantic encoder extracts important features from the data, while the channel encoder encodes those features to protect from the noise of physical environments and map it to a lower dimension, eventually saving the communication resource. The following equation defines the channel encoding process:
\vspace{-0.01in}
\begin{equation}
    X_{I}= E^{C}_{\beta}(F_{I}),
    \vspace{-0.01in}
\end{equation}
where $E^{C}_{\beta}$ represents the channel encoder with parameter $\beta$, and $X_{I}$ denotes the decoded symbol after the channel encoding process. The symbol will be transmitted through a noisy environment to the receiver since we consider the existence of noise and channel fading effect. Therefore, the received symbol at the mobile user can be expressed as follows:
\vspace{-0.01in}
\begin{equation}
    \hat{Y}= X_{I}H + N,
    \vspace{-0.01in}
\end{equation}
where $H$ denotes the fading coefficient at the receiver, and $N$ is the environment noise. With the channel gain and zero forcing detector, the transmitted signal can be estimated using the following formulation according to \cite{xie2021task}:
\vspace{-0.01in}
\begin{equation}
    \hat{X}_{I}= (H^{H}H)^{-1}H\hat{Y}= X_{I} + \hat{N},
    \vspace{-0.01in}
\end{equation}
where $\hat{X}$ is the estimated symbol after the transformed operation, and $\hat{N}$ is the environment noise. The technique converts the channel effect from multiplication to additive operation, which is easier to approximate and eliminate \cite{nguyen2023swin}. The estimated symbol is mapped back into the same dimension as the original semantic features by the channel decoder as follows:
\vspace{-0.01in}
\begin{equation}
    \hat{F}_{I}= D^{C}_{\gamma}(\hat{X}_{I}),
    \vspace{-0.01in}
\end{equation}
where $D^{C}_{\gamma}$ denotes the channel decoder with the learning parameter $\gamma$. The decoded features are then used to reconstruct the original image by the semantic decoder $D^{S}_{\sigma}$ with its learning parameter $\sigma$. The reconstructed image is acquired as follows:
\vspace{-0.01in}
\begin{equation}
    \hat{I}= D^{S}_{\sigma}(\hat{F}_{I}), \hat{I} \in R^{ 3\times H \times W}.
    \vspace{-0.01in}
\end{equation}

The objective of the image reconstruction task is to regenerate the image at the receiver site close to the original image. Therefore, the most common loss for the task is the mean square error (MSE) calculated between those images, which is expressed as:
\vspace{-0.01in}
\begin{equation}\label{lossfunction}
    L(I,\hat{I})=  \mathrm{MSE}(I,\hat{I}).
\end{equation}
\vspace{-0.4in}
\subsection{Federated Learning}
\textcolor{black}{The FL framework mainly includes four steps:\emph{ a) the server distributes a global model to clients in the network, b) clients train the received model using their local data, c) clients send the local models back to the server, d) the server aggregates model updates into a new global model.}}
\textcolor{black}{With this mechanism, FL has emerged as a proposed solution for privacy-preserving decentralized learning. In this study, we examine a real-world scenario where data is distributed among individual users and a server coordinating the overall training process.} 
\vspace{-0.1in}
\section{system model}
Semantic communication has been proven to benefit from state-of-the-art deep learning models, which can effectively perform semantic extraction and data compression. \textcolor{black}{Our study prioritizes addressing user privacy concerns within semantic communication systems rather than concentrating on the development of new deep learning techniques to enhance system performance.}
Therefore, we take advantage of the proposed deep learning models for both semantic and channel from \cite{yang2023witt}. 
\vspace{-0.2in}
\subsection{FL Framework and the Proposed FedLol}
In our proposed framework, the training process starts with mobile user $k$ receiving the model from the central server and using its data to train the model. The mobile user will update the local parameter $\boldsymbol{\Phi}_{k}$ = \{$\alpha_{k}$,$\beta_{k}$,$\gamma_{k}$,$\sigma_{k}$\} with respect to the loss defined in \eqref{lossfunction} using the following equation:
\vspace{-0.02in}
\begin{equation} 
\boldsymbol{\Phi}^r_{k} =  \boldsymbol{\Phi}_{k}^{r-1} - \eta \nabla \mathrm{L}_{k},
\vspace{-0.01in}
\end{equation}
where $r$ indicates the current local round, while $\eta$ and $\nabla\mathrm{L}_{k}$ denote the learning rate and the gradient of local loss, respectively. The model can be trained once or multiple times by the users in a single global round of the FL framework, which is denoted as the number of local epochs. Once local training is completed, each mobile user sends its local model to the server, where the global model is aggregated as follows:
\vspace{-0.02in}
\begin{equation}
    \boldsymbol{\Phi}= \sum_{k=1}^{K}\boldsymbol{\omega_{k}}\boldsymbol{\Phi}_{k},
    \vspace{-0.01in}
\end{equation}
where $\boldsymbol{\omega_{k}}$ denotes the parameter contribution of user $k$ to the global model. The $\boldsymbol{\omega}_{k}$ value depends on the FL algorithms that are being used; for example, in the FedAvg \cite{mcmahan2017communication} framework, the values are determined by $d_{k}/D$, where $d_{k}$ is the number of data possessed by user $k$ and $D$ is the total number of data in all users, respectively. In our proposed algorithm FedLol, we determine this value based on the performance of the local model. To be specific, \textcolor{black}{each local model is trained in an end-to-end manner to reconstruct the local images as closely as possible, and is associated with its own local loss. A low loss value for a particular model indicates its proficiency in the task. As a result, we prioritize that specific model by giving it more weight, rather than simply averaging the parameters from all participating models}. Therefore, the value $\boldsymbol{\omega_{k}}$ is redefined as follows:
\vspace{-0.02in}
\begin{equation}\label{weightcontribution}
    \boldsymbol{\omega_{k}}=\frac{1}{(K-1)} \frac{\sum_{k=1}^{K}(\mathrm{L}_{k})-\mathrm{L}_{k}}{\sum_{k=1}^{K}(\mathrm{L}_{k})},
    \vspace{-0.02in}
\end{equation}
where the first fraction works as a normalization term, and the second fraction makes sure the linear relationship between the update weight and the loss value. 
\begin{algorithm}[t]
   \caption{\strut Training Semantic Communication in an Efficient FL Framework: FedLol} 
   \label{alg:Alg1}
   \begin{algorithmic}[1]
       \STATE{\textbf{Initialize:} Global model $\boldsymbol{\Phi}$,
       number of global rounds $T$, local epochs $R$, update interval for channel encoder/decoder $P$.}
      \FOR{one global round t$=1,2,...,T$}
      \STATE{Check the current global round: $\textbf{if}$ t $\%$ $P$ $==1$, send the whole model $\textbf{else}$ send the semantic encoder/decoder model only.}
      \FOR{each client k$=1,2,3...,K$ $\textbf{in parallel}$}
      \STATE{Synchronize local model with the received model}
      \WHILE{client round $r$ < $R$}
      \STATE{Train the model with local data.}
      \STATE{$\boldsymbol{\Phi}_{k}^{r} \leftarrow \boldsymbol{\Phi}_{k}^{r-1}- \eta \nabla \mathrm{L}_{k}$.}
      \ENDWHILE
      \STATE{$\textbf{if}$ t $\%$ $P$ $==0$ send the whole local model \& $L_{k}$ $\textbf{else}$ send the semantic encoder/decoder models \& $L_{k}$.}    
      \ENDFOR
        \STATE{Calculate $\boldsymbol{w_{k}}$ $\forall k \in [1,K]$ as Eq.~\ref{weightcontribution} and aggregating the global model with the calculated weights.}
      \ENDFOR
      \STATE{\textbf{Output:} Global Model $\Phi$}
   \end{algorithmic}
\end{algorithm}
\vspace{-0.1in}
\subsection{Communication Efficiency for FL}
Another challenging problem with the FL framework is the communication cost, since the server and mobile users continuously exchange models in a training round. The communication overhead can be especially high when exchanging large models with a larger number of users. To address this, we study the complexity of the channel encoder/decoder and the semantic encoder/decoder. Channel encoder and decoder are only built by stacking multiple fully connected layers and a single skip connection operation. On the other hand, the semantic encoder and decoder are constructed using a computationally intensive Swin Transformer, known for its intricate architecture in Fig.~\ref{systemmodel}. Given this analysis, we consider partial transmission of these modules within a global training round. Specifically, the semantic encoder/decoder will be transmitted in every round, while the channel encoder/decoder parameter will be aligned after certain global rounds. The details of the proposed method can be found in Algorithm~\ref{alg:Alg1}.
\vspace{-0.1in}
\section{simulation results}
In this section, we first present the simulation settings for the FL environment, the architecture and size of semantic and channel encoder/decoder, and finally, the improvement in the performance of the proposed framework. The FL setting can be found in Tab.~\ref{tab1}.
\vspace{-0.15in}
\subsection{Model and Dataset}
\emph{Semantic encoder/decoder model}: We utilize the Swin transformer architecture from \cite{liu2021swin}, as shown in Fig.~\ref{systemmodel}. However, to adapt the model for mobile users, we reduce the number of transformer layers to two layers per stage, and we have a total of 4 stages. Two layers is the lowest number per stage allowed by its design, according to its original study. The model sizes of the semantic encoder and decoder are 55.12 MB and 53.41 MB, respectively. 

\emph{Channel encoder/decoder model}: Its architecture is quite simple, comprising a set of seven fully connected layers and a skip connection operation; the signal-to-noise ratio (SNR) information will be injected by a set of dense layers into the middle of two consecutive fully connected (FC) layers as in \cite{yang2023witt}. The size of the channel encoder/decoder is 25.07 MB. 

\begin{figure*}[t!]
	\centering
	\subfigure[]{\includegraphics[width=0.33\linewidth]{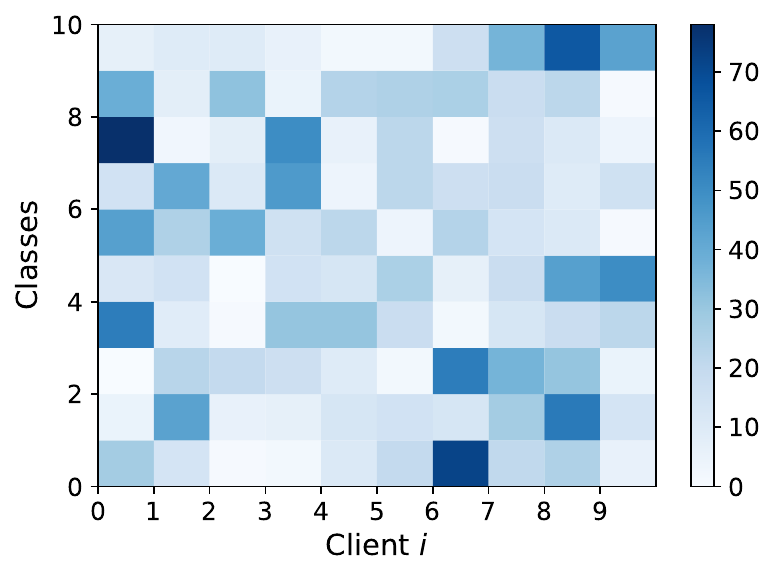}\label{fig2a}}
	\hfil
	\centering
	\subfigure[]{\includegraphics[width=0.33\linewidth]{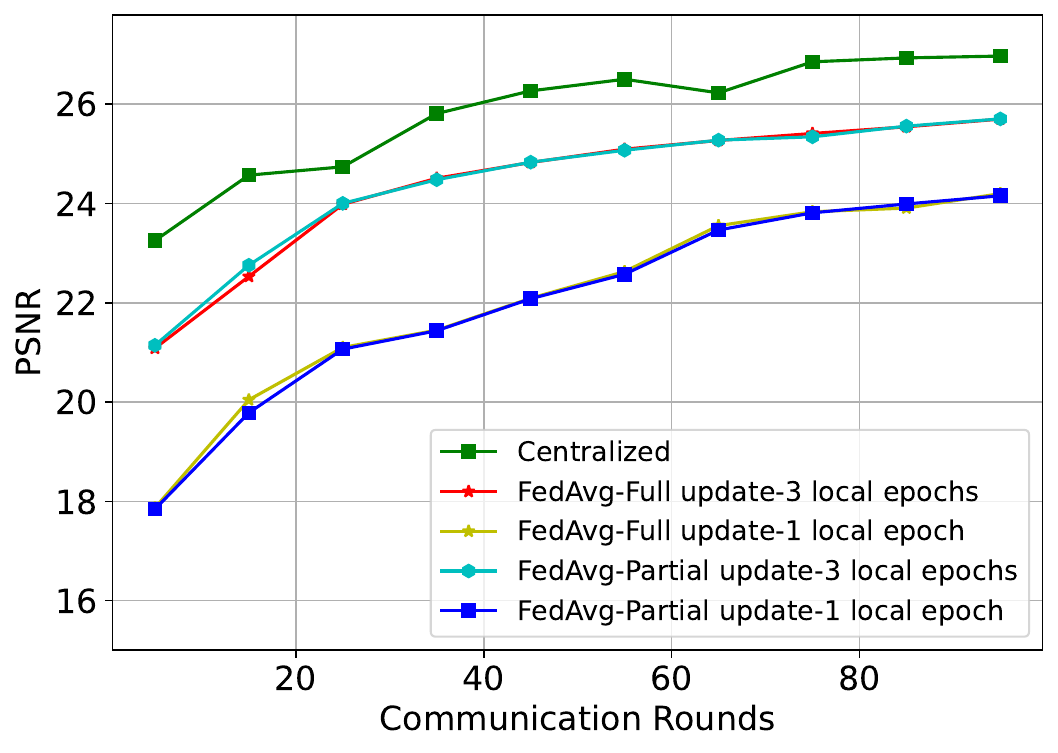}\label{fig2b}}
	\hfil
	\centering
 	\subfigure[]{\includegraphics[width=0.173\linewidth]{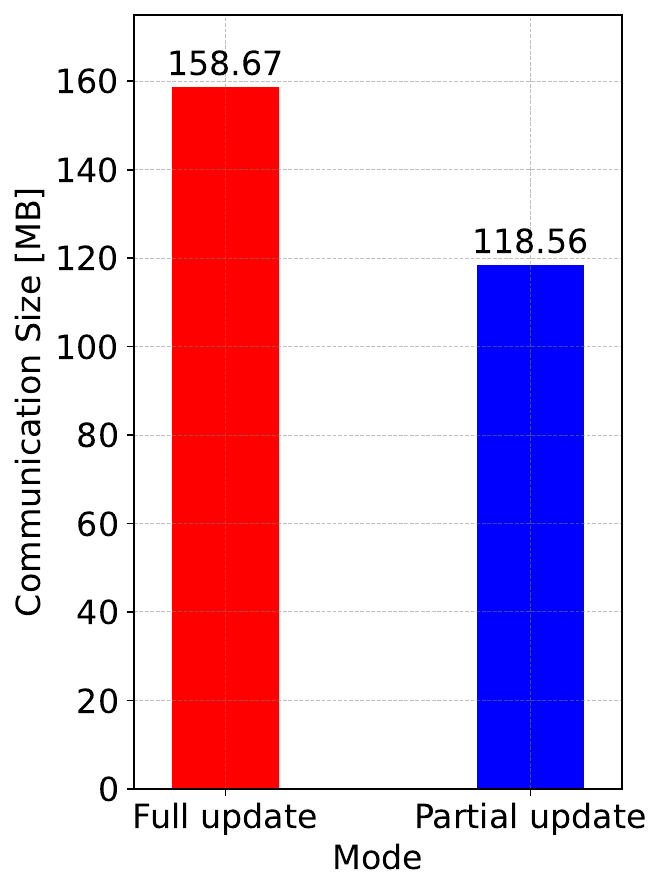}\label{fig2c}}
	\hfil
	\caption{\textcolor{black}{(a) The data distribution of each client using non-IID data partition. The color bar indicates the number of data samples, while each rectangle points out the number of data samples of a specific class in a client. (b) The convergence of the FedAvg in terms of full-update and partial-update. (c) Communication cost of the full-update and partial-update scenarios.}}
    \label{Metric}
\end{figure*}

\emph{ImageNet10 dataset}: \textcolor{black}{For the training process, we use a subset of 10 classes from the ImageNet dataset \cite{russakovsky2015imagenet}. We select around 200 images for each class, omitting images with dimensions lower than 400 × 400. We take advantage of the Dirichlet distribution to generate non-IID datasets for each client. The local data distribution of each client can be found in Fig.~g\ref{fig2a}.}

\emph{DIV2K dataset}: \textcolor{black}{To evaluate the performance of the trained models, we use the 2K resolution images from the test set of DIV2K dataset \cite{agustsson2017ntire}, which is widely accepted in semantic communication literature to evaluate the system's performance.}

\vspace{-0.1in}
\begin{table}[t!]
	\caption{Simulation Parameters}
	\textbf{\label{tab:table_simulation1}} 
	\renewcommand\arraystretch{1}
	\begin{center}
		\begin{tabular}{|p{2.8cm}|p{0.7cm}||p{2.8cm}|p{0.7cm}|}
			\hline
			\hfil \textbf{Parameter} & \hfil \textbf{Value} & \hfil \textbf{Parameter} & \hfil \textbf{Value} \\ \hline \hline
			\ \hfil Number of clients & 10  & \hfil The compression ratio &  1/16 \\ \hline
			\ \hfil Learning rate & 1$e^{-4}$ & \hfil  Communication round & 100 \\ \hline
   			\ \hfil Batch Size & 16 & \hfil Local training rounds $R$ & 3 \\ \hline
		\end{tabular}
		\label{tab1}
	\end{center}
\end{table}

\subsection{Benchmark}
To fully evaluate the effectiveness of each proposal mechanism, we consider a comprehensive of benchmarks.
\begin{itemize}
    \item \emph{Centralized Training:} All the images are considered to be collected at the centralized server for the training process. This benchmark serves as an upper bound of the FL framework.
    \item \emph{FedAvg-Full \&  Partial update:} \textcolor{black}{All and partial parameters from mobile users will be transmitted to the server for aggregation in each round. The weight contribution to the aggregation process is the ratio between the number of samples of that client and the total samples of all clients.}
    \item \emph{FedProx:} \textcolor{black}{It is considered as a generalization and re-parametrization of FedAvg to solve the heterogeneity problem in federated networks.}
    \item \emph{MOON:} \textcolor{black}{It is specifically tailored to optimize performance on image datasets by leveraging the similarity between model representations to refine the local training of individual parties.}
\end{itemize}
\subsection{Evaluating Communication Efficiency of FL.}
\textcolor{black}{In this sub-section, we focus on the system's performance when it is partially updated; therefore, we use the same model aggregation approach as in the FegAvg framework. In Fig.\ref{fig2b}, we compare the performance of the proposed method-partial update with the full update and centralized training in Peak Signal to Noise Ratio (PSNR) value as in \cite{bourtsoulatze2019deep}. It is easy to recognize the performance of the centralized scheme is the highest. However, it is difficult to achieve it in practical scenarios due to the data privacy of users. In the meantime, the performance of FL improves slowly but steadily through each training round. Notably, the PSNR metric values remain almost indistinguishable for both partial and full updates.} This result proves our assumption that the channel encoder/decoder with simple architecture can achieve good performance, and its parameter values slowly change. In contrast, the complex design of the semantic encoder/decoder requires a long time to improve performance. Thus, their parameter values are sent to the mobile user and aggregated in every communication epoch. Fig.~\ref{fig2c} shows the average transmission size in one communication round; the partial update technique decreases the amount of transmitted data by 25.28\%. In addition, it is worth noticing that the increase in local training rounds boosts the training process to converge faster. 
\vspace{-0.1in}
\begin{figure}[t!]
	\centering
	\subfigure[]{\includegraphics[width=0.49\linewidth]{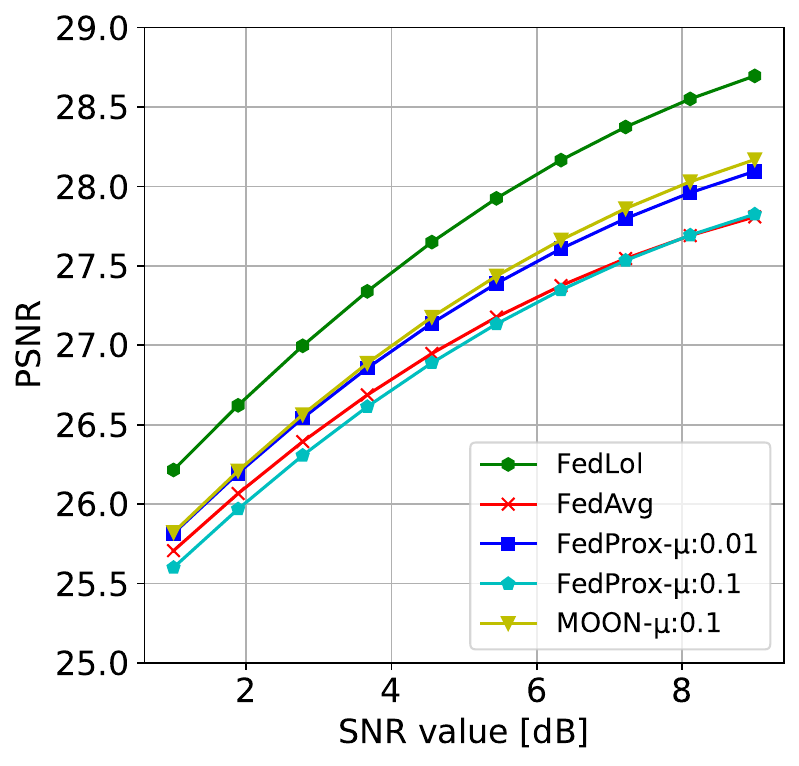}\label{Metrica}}
	\hfil
	\subfigure[]{\includegraphics[width=0.49\linewidth]{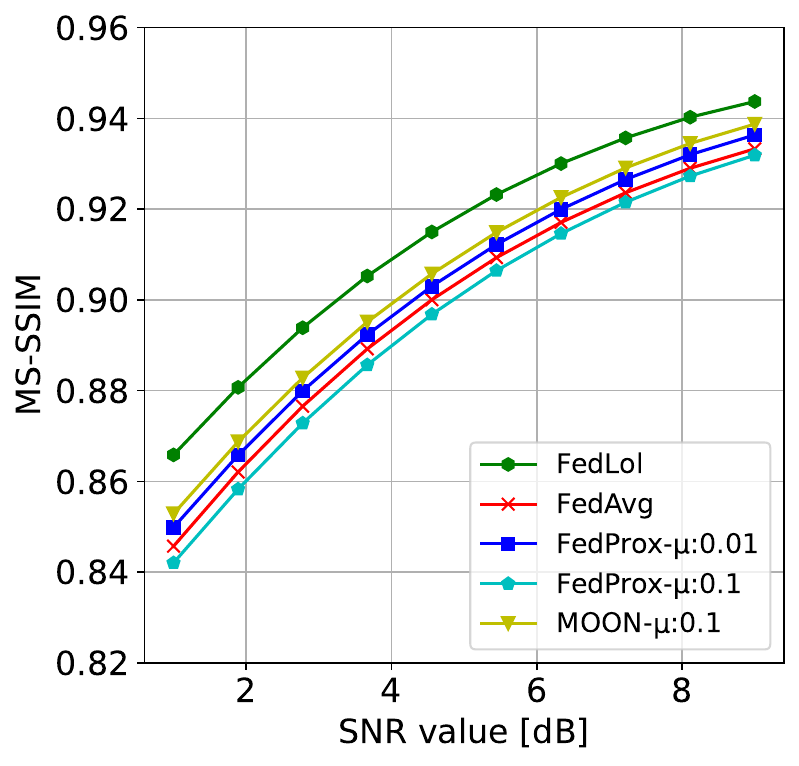}\label{Metricb}}
	\hfil
	\caption{\textcolor{black}{The PSNR (a) and MS-SSIM (b) values of the proposed algorithm compared to other benchmarks.}}
\end{figure}
\vspace{-0.12in}
\subsection{Evaluating the Proposed Aggregation Approach}
In this subsection, we evaluate the efficiency of the proposed method FedLol with FedAvg, FedProx, and MOON benchmarks in the PSNR metric and the MS-SSIM metric \cite{wang2003multiscale}, which is a popular metric for evaluating the quality of the reconstructed image. As shown in Fig.~\ref{Metrica}, when the SNR value increases, signifying a decrease in noise, the performance of all algorithms achieves higher value in both metrics. \textcolor{black}{Overall, our aggregation technique achieves the highest value in two metrics and secures a big performance gap compared to other benchmark schemes. Specifically, the proposal mechanism achieves 26.215 dB when the SNR is at 1 dB, while the MOON only acquires 25.822 dB, and this gap keeps widening with the larger value of SNR. While the performance of FedProx heavily depends on the value of the hyper-parameter $\mu$, its performance can get closer to MOON with the right value and perform even worse than FedAvg when $\mu=$ 0.1. In Fig.~\ref{Metricb}, while the performance gaps may not appear significant at first glance, they are indeed distinguishable due to the fact that the maximum value of MS-SSIM is 1, and our proposed mechanism continues to surpass other benchmarks by a substantial margin.}

\vspace{-0.18in}
\section{Conclusion}
{\color{black}In this letter, we leverage the federated learning framework to exploit private user data for training a semantic communication system.} \textcolor{black}{In addition, we reduce the communication overhead in the training process by limiting the {\color{black}frequency of aggregation and broadcasting} the channel encoder and decoder}. Due to their simple architecture, they can still be combined with the semantic encoder/decoder and produce {\color{black}comparable} results with the full update scenario, while alleviating up to 25\% of transmitting data. Besides, we propose a new simple aggregation mechanism, FedLol, which {\color{black}aggregates the global model towards local models associated with low loss value.} The proposed method outperforms the FedAvg, FedProx, and MOON in both metrics without additional modules or communication costs. \textcolor{black} 
{These results proved our assumptions: {\color{black}firstly}, not all the parameters need to be transmitted and aggregated in one global training round for semantic communication; secondly, the performance of individual clients significantly influences the global model aggregation in the context of federated learning for image reconstruction tasks.}
\bibliographystyle{IEEEtran}
\vspace{-0.2in}
\bibliography{References}
\end{document}